\documentclass{article}

\usepackage{ijcai13}

\newtheorem {definition}{Definition}

\newtheorem {theorem}{Theorem}

\newtheorem {example}{Example}

\usepackage{amsmath}
\usepackage{amssymb}
\usepackage{rotating}

\begin{document}

\title{Logical Probability Preferences}

\author{ Emad Saad \\
emsaad@gmail.com
}

\maketitle

\begin{abstract}

We present a unified logical framework for representing and reasoning about both probability quantitative and qualitative preferences in probability answer set programming \cite{Saad_NHPP,Saad_EHPP,Saad_DHPP}, called probability answer set optimization programs. The proposed framework is vital to allow defining probability quantitative preferences over the possible outcomes of qualitative preferences. We show the application of probability answer set optimization programs to a variant of the well-known nurse restoring problem \cite{Nurse}, called {\em the nurse restoring with probability preferences problem}. To the best of our knowledge, this development is the first to consider a logical framework for reasoning about probability quantitative preferences, in general, and reasoning about both probability quantitative and qualitative preferences in particular.
\end{abstract}

\section{Introduction}

Probabilistic reasoning is inevitable in almost all real-world applications. Therefore, developing well-defined frameworks for representing and reasoning in the presence of probabilistic knowledge and under probabilistic environments is vital. Thus, many frameworks have been developed for representing and reasoning in the presence of probabilistic knowledge and under probabilistic environments. Among these frameworks are probability answer set programming which are probability logic programs with probability answer set semantics \cite{Saad_NHPP,Saad_EHPP,Saad_DHPP}.

The importance of the probability answer set programming frameworks of \cite{Saad_NHPP,Saad_EHPP,Saad_DHPP} lies in the fact that the probability answer set programming frameworks of \cite{Saad_NHPP,Saad_EHPP,Saad_DHPP} have been shown applicable to a variety of fundamental probabilistic reasoning tasks. These probabilistic reasoning tasks include, but are not limited to, probabilistic planning \cite{SaadPlan}, probabilistic planning with imperfect sensing actions \cite{Saad_Sensing}, reinforcement learning in MDP environments \cite{Saad_MDP}, reinforcement learning in POMDP environments \cite{Saad_Learn_Sense}, and Bayes reasoning \cite{Saad_EHPP}. Moreover, in \cite{SaadSSAT} it has been proved that stochastic satisfiability (SSAT) can be modularly encoded as probability answer set programs with probability answer set semantics, therefore, the applicability of SSAT to variety of fundamental probabilistic reasoning tasks also carry over to probability answer set programming \cite{Saad_NHPP,Saad_EHPP,Saad_DHPP}.

In addition, the probability answer set programming frameworks of \cite{Saad_NHPP,Saad_EHPP,Saad_DHPP} are strictly expressive. This is because the way how a rule fires in \cite{Saad_NHPP,Saad_EHPP,Saad_DHPP} is close to the way how it fires in classical answer set programming \cite{Gelfond_A,Gelfond_B}, which makes any possible extension to \cite{Saad_NHPP,Saad_EHPP,Saad_DHPP} to more expressive forms of probability answer set programming is more flexible and more intuitive.

Extended and normal disjunctive hybrid probability logic programs with probability answer set semantics is an expressive probability answer set programming framework \cite{Saad_DHPP} that generalize and subsume extended hybrid probability logic programs \cite{Saad_EHPP} and normal hybrid probability logic programs \cite{Saad_NHPP} with probability answer set semantics as well as classical extended and classical normal disjunctive logic programs with classical answer set semantics \cite{Gelfond_B} in a unified logical framework to allow non-monotonic negation, classical negation, and disjunctions under probabilistic uncertainty.

The probability answer set programming framework of \cite{Saad_DHPP} allows directly and intuitively to represent and reason in the presence of both probabilistic uncertainty and qualitative uncertainty in a unified logical framework. This is necessary to provide the ability to assign probabilistic uncertainly over the possible outcomes of qualitative uncertainty, which is required in most real life applications, e.g., representing and reasoning about {\em probability quantitative preferences}. However, the probability answer set programming framework of \cite{Saad_DHPP} is insufficient for representing and reasoning about probability quantitative preferences. This is because any probability answer set program encoding of a probability quantitative preferences reasoning problem provides all possible solutions to the problem that {\em satisfy} the probability quantitative preferences represented in the probability answer set program encoding of the problem, rather than {\em ranking} all the possible solutions that satisfy these probability quantitative preferences from the top preferred solution to the least preferred solution.

For example, consider the following simple instance of the well-known Nurse Restoring Problem from Operation Research \cite{Nurse}. Consider that a nurses, $a$, in a hospital, need to be assigned to one shift among two shifts $s_1,s_2$ in a given day, $d$, such that nurse $a$ is assigned exactly one shift. If nurse $a$ is neutral regarding servicing at either shifts in that given day, then classical disjunctive logic program can be used to model this problem as a classical disjunctive logic program of the form
\[
service(a, s_1,d)  \; \vee \; service(a, s_2,d)
\]
with $\{ service(a, s_1,d) \}$ and $\{ service(a, s_2,d)  \}$ are the possible classical answer sets, according to the classical answer set semantics of classical disjunctive logic programs \cite{Gelfond_B}. Consider that nurse, $a$, prefers to service at shift $s_1$ over shift $s_2$ in day, $d$, due to some circumstances, where this preference relation is specified as a probability distribution over the shifts $s_1, s_2$ in the day $d$. Consider also that the probability nurse $a$ prefers to service at shift $s_1$ in day $d$ is characterized by the probability value $0.7$ and the probability nurse $a$ prefers to service at shift $s_2$ in day $d$ is characterized by the probability value $0.4$. In this case, classical disjunctive logic programs cannot represent the nurse's preferences over the shifts in the day $d$, since classical disjunctive logic programs are incapable in general of reasoning in the presence of probabilistic uncertainty. However, this variant of the nurse restoring problem can be intuitively represented as disjunctive hybrid probability logic program with probability answer set semantics of the form
\[
service(a, s_1,d):0.7  \; \vee \; service(a, s_2,d):0.4
\]
We call this variant of the nurse restoring problem {\em Nurse Restoring with probability Preferences problem}.

The probability answer set program encoding of the nurse restoring with probability preferences problem instance described above has two probability answer sets namely $\{ service(a, s_1,d):0.7 \}$ and $\{ service(a, s_2,d):0.4 \}$, according to the probability answer set semantics of probability answer set programming of \cite{Saad_DHPP}. It is clear that the probability answer set $\{ service(a, s_1,d):0.7 \}$ represents nurse $a$'s top servicing preferences, which means that the probability answer set $\{ service(a, s_1,d):0.7 \}$ is the most preferred probability answer set according to the {\em probability quantitative preferences} represented by the probability answer set program. Furthermore, assume that nurse $a$ is neutral regarding servicing at shifts $s_1$ and $s_2$, where this servicing preference of nurse $a$ is characterized by the probability value $0.2$ for both shifts. In this case, this nurse restoring with probability preferences problem instance can be represented as a probability answer set program of the form \[service(a, s_1,d):0.2 \; \vee \; service(a, s_2,d):0.2\] with $\{service(a, s_1,d):0.2 \}$ and $\{ service(a, s_2,d):0.2 \}$ are the probability answer sets, according to the probability answer set semantics of probability answer set programming of \cite{Saad_DHPP}. Although nurse $a$ is neutral regarding servicing at either shifts with probability preference $0.2$ each, however, it can be the case that nurse $a$ has more appeal in servicing at shift $s_1$ over shift $s_2$ ({\em qualitative preferences}). This makes $\{ service(a, s_1,d):0.2 \}$ is the most preferred probability answer set in this case.

The existing probability answer set programs semantics \cite{Saad_NHPP,Saad_EHPP,Saad_DHPP} does not have the ability to rank probability answer sets either according to probability quantitative preferences or according to qualitative preferences. Rather, probability answer set programs semantics is capable of finding probability answer sets that {\em satisfy} probability quantitative preferences represented by the probability answer set program and considers all the resulting probability answer sets as equally preferred. Although, in many applications, it is necessary to rank the probability answer sets generated by the probability answer set programs from the top (most) preferred probability answer set to the least preferred probability answer set, where the top (most) preferred probability answer set is the one that is most desirable. This requires probability answer set programs to be capable of representing both probability quantitative and qualitative preferences and to be capable of reasoning in the presence of both probability quantitative and qualitative preferences across probability answer sets.
\\
\\
In this paper we develop a unified logical framework that is capable of representing and reasoning about both probability quantitative and qualitative preferences. This is accomplished by defining the notion of {\em probability answer set optimization programs}. A probability answer set optimization program is a set of probability logic rules under the probability answer set semantics whose probability answer sets are ranked according to probability preferences relations specified by the user. Probability answer set optimization programs modify and generalize the classical answer set optimization programs described in \cite{ASO}. We show the application of probability answer set optimization programs to a variant of the well-known nurse restoring problem, called {\em the nurse restoring with probability preferences problem}, where a probability answer set program (disjunctive hybrid probability logic program with probability answer set semantics) \cite{Saad_DHPP} is used as probability answer sets generator rules. To the best of our knowledge, this development is the first to consider a logical framework for reasoning about probability quantitative preferences, in general, and reasoning about both probability quantitative and qualitative preferences in particular.

\section{Probability Answer Sets}

We use probability logic rules under the probability answer set semantics to generate probability answer sets, that are ultimately ranked by probability preference rules. Therefore, in this section we review the probability answer set semantics of disjunctive hybrid probability logic sets of rules, a form of probability answer set programming, as described in \cite{Saad_DHPP}.

\subsection{Syntax}

Let $\cal L$ denotes an arbitrary first-order language with finitely many predicate symbols, function symbols, constants, and infinitely many variables. A standard atom is a predicate in $\cal {B_L}$, where $\cal {B_L}$ is the Herbrand base of $\cal L$. Non-monotonic negation or the negation as failure is denoted by $not$. In disjunctive hybrid probability logic rules, probabilities are assigned to primitive events (atoms) and compound events (conjunctions or disjunctions of atoms) as intervals in $C[0,1]$, where $C[0, 1]$ denotes the set of all closed intervals in $[0, 1]$. For $[\alpha_1, \beta_1], [\alpha_2, \beta_2] \in C[0, 1]$, the \emph{truth order} $\leq_t$ on $C[0, 1]$ is defined as $[\alpha_1, \beta_1] \leq_t [\alpha_2, \beta_2]$ iff $\alpha_1 \leq \alpha_2$ and $\beta_1 \leq \beta_2$.

The type of dependency among the primitive events within a compound event is described by \emph{a probabilistic strategy}, which can be a \emph{conjunctive} p-strategy or a \emph{disjunctive} p-strategy. Conjunctive (disjunctive) p-strategies are used to combine events belonging to a conjunctive (disjunctive) formula \cite{Saad_NHPP}. The \emph{probabilistic composition function}, $c_\rho$, of a probabilistic strategy (p-strategy), $\rho$, is a mapping $c_\rho : C[0,1] \times C[0, 1] \rightarrow C[0, 1]$, where the probabilistic composition function, $c_\rho$, computes the probability interval of a conjunction (disjunction) of two events from the probability of its components. Let $M = \{\!\!\{[\alpha_1, \beta_1], \ldots, [\alpha_n, \beta_n]\}\!\!\}$ be a multiset of probability intervals. For convenience, we use $c_\rho M$ to denote $c_\rho ([\alpha_1, \beta_1], c_\rho ([\alpha_2, \beta_2],\ldots, c_\rho([\alpha_{n-1}, \beta_{n-1}],[\alpha_n, \beta_n]))\ldots )$.

A \emph{probability annotation} is a probability interval of the form $[\alpha_1, \alpha_2]$, where $\alpha_1, \alpha_2$ are called probability annotation items. A \emph{probability annotation item} is either a constant in $[0, 1]$ (called \emph{probability annotation constant}), a variable ranging over $[0, 1]$ (called \emph{probability annotation variable}), or $f(\alpha_1,\ldots,\alpha_n)$ (called \emph{probability annotation function}), where $f$ is a representation of a computable function $f: ([0, 1])^n \rightarrow [0, 1]$ and $\alpha_1,\ldots,\alpha_n$ are probability annotation items.

Let $S = S_{conj} {\cup} S_{disj}$ be an arbitrary set of p-strategies, where $S_{conj}$ ($S_{disj}$) is the set of all conjunctive (disjunctive) p-strategies in $S$. A \emph{hybrid basic formula} is an expression of the form  $a_1 \wedge_\rho \ldots \wedge_\rho a_n$ or $a_1 \vee_{\rho'} \ldots \vee_{\rho'} a_n$, where $a_1, \ldots, a_n$ are atoms and $\rho$ and $\rho'$ are p-strategies. Let $bf_S({\cal B_L})$ be the set of all ground hybrid basic formulae formed using distinct atoms from ${\cal B_L}$ and p-strategies from $S$. If $A$ is a hybrid basic formula and $\mu$ is a probability annotation then $A:\mu$ is called a probability annotated hybrid basic formula. A disjunctive hybrid probability logic rule is an expression of the form
\begin{eqnarray}
a_1:\mu_1 \vee \ldots \vee a_k:\mu_k \leftarrow A_{k+1}:\mu_{k+1}, \ldots, A_m:\mu_m, \notag \\
not\; A_{m+1}:\mu_{m+1},\ldots, not\;A_{n}:\mu_{n}, \label{rule:dhpl}
\end{eqnarray}
where $a_i$ ($1 \leq i \leq k$) are atoms, $A_i$ ($k+1 \leq i \leq n$) are hybrid basic formulae, and $\mu_i$ ($1 \leq i \leq n$) are probability annotations. A disjunctive hybrid probability logic rule says that if for each $A_i:\mu_i$, where $k+1 \leq i \leq m$, the probability interval of $A_i$ is at least $\mu_i$ and for each $not\; A_j:\mu_j$, where $m+1 \leq j \leq n$, it is \emph{not believable} that the probability interval of $A_j$ is at least $\mu_j$, then there exist at least $a_i$, where $1 \leq i \leq k$, such that the probability interval of $a_i$ is at least $\mu_i$. Associated with every set of disjunctive hybrid probability logic rules is a mapping, $\tau$, where $\tau: {\cal B_L} \rightarrow S_{disj}$. The mapping $\tau$ associates to each atom $a$ in ${\cal B_L}$ a disjunctive p-strategy that is used to combine the probability intervals obtained from different disjunctive hybrid probability logic rules with the atom, $a$, appearing in their heads.

A disjunctive hybrid probability logic rule is ground if it does not contain any variables. For the simplicity of the presentation, hybrid basic formulae that appearing in a disjunctive hybrid probability logic rule without probability annotations are assumed to be associated with the annotation $[1,1]$. In addition, annotated hybrid basic formulae of the form $A:[\alpha, \alpha]$ are simply presented as $A:\alpha$.

\subsection {Probability Answer Set Semantics}

A probabilistic interpretation (p-interpretation), $h$, for a set of disjunctive hybrid probability logic rules is a mapping $h: bf_{S}({\cal B_L}) \rightarrow C[0, 1]$. Let $r$ be a disjunctive hybrid probability logic rule of form (\ref{rule:dhpl}) and $head(r) = a_1:\mu_1 \vee \ldots \vee a_k:\mu_k$ and $body(r) = A_{k+1}:\mu_{k+1}, \ldots, A_m:\mu_m, not\; A_{m+1}:\mu_{m+1},\ldots, not\;A_{n}:\mu_{n}$.

\begin{definition}
Let $R$ be a set of ground disjunctive hybrid probability logic rules, $\tau$ be a mapping $\tau: {\cal B_L} \rightarrow S_{disj}$ associated to $R$, $h$ be a p-interpretation for $R$, and $r$ be a disjunctive hybrid probability logic rule of the form (\ref{rule:dhpl}). Then:

\begin{enumerate}

\item $h$ satisfies $a_i:\mu_i$ in $head(r)$ iff  $\mu_i \leq_t h(a_i)$.

\item $h$ satisfies $A_i : \mu_i$ in $body(r)$ iff  $\mu_i \leq_t h(A_i)$.

\item $h$ satisfies $not\;A_j:\mu_j$ in $body(r)$ iff $\mu_j \nleq_t h(A_j)$.

\item $h$ satisfies $body(r)$ iff $\forall(k+1 \leq i \leq m), h$ satisfies $A_i : \mu_i$ and $\forall(m+1 \leq j
\leq n), h$ satisfies $not\;A_j : \mu_j$.

\item $h$ satisfies $head(r)$ iff $\exists i$ $(1 \leq i \leq k)$ such that $h$ satisfies $a_i : \mu_i$.

\item $h$ satisfies $r$ iff $h$ satisfies $head(r)$ whenever $h$ satisfies $body(r)$ or $h$ does not satisfy $body(r)$.

\item $h$ satisfies $R$ iff $h$ satisfies every disjunctive hybrid probability logic rule in $R$ and

\begin{itemize}
\item $c_{\tau(a_i)} \{\!\!\{\mu_i \; | \; head(r) \leftarrow body(r) \in R\}\!\!\}\leq_t h(a_i)$ such that $h$ satisfies $body(r)$ and $h$ satisfies $a_i:\mu_i$ in the $head(r)$.

\item $c_\rho \{\!\!\{ h(a_1), \ldots, h(a_n) \}\!\!\} \leq_t h(A)$ such that $a_1, \ldots, a_n$ are atoms in $\cal B_L$ and $A = a_1 *_\rho \ldots *_\rho a_n$, where $* \in \{\vee, \wedge\}$.
\end{itemize}

\end{enumerate}
\end{definition}
A probabilistic model (\emph{p-model}) of a set of disjunctive hybrid probability logic rules, $R$, associated with a mapping $\tau: {\cal B_L} \rightarrow S_{disj}$, is a p-interpretation for $R$ that satisfies $R$. A p-model $h$ of $R$ is \emph{minimal} w.r.t. $\leq_t$ iff there does not exist a p-model $h'$ of $R$ such that $h' <_t h$. Let $R$ be a set of ground disjunctive hybrid probability logic rules, $\tau$ be a mapping $\tau: {\cal B_L} \rightarrow S_{disj}$ associated to $R$, and $h$ be a p-interpretation for $R$. Then, the probabilistic reduct, $R^h$, of $R$ w.r.t. $h$ is the set of ground non-monotonic-negation-free disjunctive hybrid probability logic rules associated to $\tau$ and

\begin{eqnarray*}
a_1 : \mu_1 \; \vee \ldots \; \vee \; a_k:\mu_k \leftarrow A_{k+1}:\mu_{k+1}, \ldots, A_m : \mu_m
\end{eqnarray*}
is in $R^h$ iff
\begin{eqnarray*}
a_1:\mu_1 \vee \ldots \vee a_k:\mu_k \leftarrow A_{k+1}:\mu_{k+1}, \ldots, A_m:\mu_m, \\ not\; A_{m+1}:\mu_{m+1},\ldots, not\;A_{n}:\mu_{n}
\end{eqnarray*}
is in $R$ and $\forall (m+1 \leq j \leq n),\:  \mu_j \nleq_t h(A_j)$.

\begin{definition} A p-interpretation, $h$, for a set of ground disjunctive hybrid probability logic rules, $R$, associated to a mapping $\tau: {\cal B_L} \rightarrow S_{disj}$, is a probabilistic answer set for $R$ if $h$ is $\leq_t$-minimal p-model for $R^h$.
\end{definition}

\section{Probability Answer Set Optimization Programs}

Probability answer set optimization programs are probability logic programs under the probability answer set semantics whose probability answer sets are ranked according to probability preference rules represented in the programs. A probability answer set optimization program, $\Pi$, is a pair of the form \\ $\Pi = \langle R_{gen} \cup R_{pref}, \tau \rangle$, where $R_{gen} \cup R_{pref}$ is a union of two sets of probability logic rules and $\tau$ is a mapping, $\tau: {\cal B_L} \rightarrow S_{disj}$, associated to the set of probability logic rules $R_{gen}$. The first set of probability logic rules, $R_{gen}$, is called the generator rules that generate the probability answer sets that satisfy every probability logic rule in $R_{gen}$ and the mapping $\tau$ associates to each atom, $a$, appearing in $R_{gen}$, a disjunctive p-strategy that is used to combine the probability intervals obtained from different probability logic rules in $R_{gen}$ with an atom $a$ appearing in their heads. $R_{gen}$ is any set of probability logic rules with well-defined probability answer set semantics including normal, extended, and disjunctive hybrid probability logic rules \cite{Saad_NHPP,Saad_EHPP,Saad_DHPP}, as well as hybrid probability logic rules with probability aggregates (all are forms of {\em probability answer set programming}).

The second set of probability logic rules, $R_{pref}$, is called the {\em probability preference rules}, which are probability logic rules that represent the user's {\em probability quantitative} and {\em qualitative preferences} over the probability answer sets generated by $R_{gen}$. The probability preference rules in $R_{pref}$ are used to rank the generated probability answer sets from $R_{gen}$ from the top preferred probability answer set to the least preferred probability answer set. Similar to \cite{ASO}, an advantage of probability answer set optimization programs is that $R_{gen}$ and $R_{pref}$ are independent. This makes probability preference elicitation easier and the whole approach is more intuitive and easy to use in practice.

In our introduction of probability answer set optimization programs, we focus on the syntax and semantics of the {\em probability preference rules}, $R_{pref}$, of the probability answer set optimization programs, since the syntax and semantics of the probability answer sets generator rules, $R_{gen}$, are the same as syntax and semantics of any set of probability logic rules with well-defined probability answer set semantics as described in \cite{Saad_NHPP,Saad_EHPP,Saad_DHPP}.

\subsection{Probability Preference Rules Syntax}

Let ${\cal L}$ be a first-order language with finitely many predicate symbols, function symbols, constants, and
infinitely many variables. A literal is either an atom $a$ in ${\cal B_L}$ or the negation of an atom $a$ ($\neg a$), where ${\cal B_L}$ is the Herbrand base of ${\cal L}$ and $\neg$ is the classical negation. Non-monotonic negation or the negation as failure is denoted by $not$. Let $Lit$ be the set of all literals in ${\cal L}$, where $Lit = \{a \:| \: a \in {\cal B_L} \} \cup \{\neg a \: | \: a \in {\cal B_L}\}$. A \emph{probability annotation} is a probability interval of the form $[\alpha_1, \alpha_2]$, where $\alpha_1, \alpha_2$ are called probability annotation items. A \emph{probability annotation item} is either a constant in $[0, 1]$ (called {\em probability annotation constant}), a variable ranging over $[0, 1]$ (called \emph{probability annotation variable}), or $f(\alpha_1,\ldots,\alpha_n)$ (called \emph{probability annotation function}) where $f$ is a representation of a computable function $f: ([0, 1])^n \rightarrow [0, 1]$ and $\alpha_1,\ldots,\alpha_n$ are probability annotation items.

Let $S = S_{conj} {\cup} S_{disj}$ be an arbitrary set of p-strategies, where $S_{conj}$ ($S_{disj}$) is the set of all conjunctive (disjunctive) p-strategies in $S$. A \emph{hybrid literals} is an expression of the form  $l_1 \wedge_\rho \ldots \wedge_\rho l_n$ or $l_1 \vee_{\rho'} \ldots \vee_{\rho'} l_n$, where $l_1, \ldots, l_n$ are literals and $\rho$ and $\rho'$ are p-strategies from $S$. $bf_S(Lit)$ is the set of all ground hybrid literals formed using distinct literals from $Lit$ and p-strategies from $S$. If $L$ is a hybrid literal $\mu$ is a probability annotation then $L:\mu$ is called a probability annotated hybrid literal. Let ${\cal A}$ be a set of probability annotated hybrid literals. A boolean combination over ${\cal A}$ is a boolean formula over probability annotated hybrid literals in ${\cal A}$ constructed by conjunction, disjunction, and non-monotonic negation ($not$), where non-monotonic negation is combined only with probability annotated hybrid literals.

\begin{definition} A probability preference rule, $r$, over a set of probability annotated hybrid literals, ${\cal A}$, is an expression of the form
\begin{eqnarray}
C_1 \succ  C_2 \succ \ldots \succ C_k \leftarrow L_{k+1}:\mu_{k+1},\ldots, L_m:\mu_m, \notag \\
not\; L_{m+1}:\mu_{m+1},\ldots, not\;L_{n}:\mu_{n}
\label{rule:pref}
\end{eqnarray}
where $L_{k+1}:\mu_{k+1}, \ldots, L_{n}:\mu_{n}$ are probability annotated hybrid literals and $C_1, C_2, \ldots, C_k$ are boolean combinations over ${\cal A}$.
\end{definition}
Let $body(r) = L_{k+1}:\mu_{k+1},\ldots, L_m:\mu_m, not\; L_{m+1}:\mu_{m+1},\ldots, not\;L_{n}:\mu_{n}$ and $head(r) = C_1 \succ C_2 \succ \ldots \succ C_k$, where $r$ is a probability preference rule of the form (\ref{rule:pref}). Intuitively, a probability preference rule, $r$, of the form (\ref{rule:pref}) means that any probability answer set that satisfies $body(r)$ and $C_1$ is preferred over the probability answer sets that satisfy $body(r)$, some $C_i$ $(2 \leq i \leq k)$, but not $C_1$, and any probability answer set that satisfies $body(r)$ and $C_2$ is preferred over probability answer sets that satisfy $body(r)$, some $C_i$ $(3 \leq i \leq k)$, but neither $C_1$ nor $C_2$, etc.

\begin{definition} A probability answer set optimization program, $\Pi$, is a pair of the form $\Pi = \langle R_{gen} \cup R_{pref}, \tau \rangle$, where $R_{gen}$ is a set of probability logic rules with well-defined probability answer set semantics, the {\em generator} rules, $R_{pref}$ is a set of probability preference rules, and $\tau$ is the mapping $\tau: {\cal B_L} \rightarrow S_{disj}$ that associates to each atom, $a$, appearing in $R_{gen}$ a disjunctive p-strategy. 
\end{definition}

\subsection{Probability Preference Rules Semantics}

In this section, we define the satisfaction of probability preference rules and the ranking of the probability answer sets with respect to a probability preference rule and with respect to a set of probability preference rules. We say that a probability preference rule is ground if it does not contain any variables. A probability answer set optimization program, $\Pi = \langle R_{gen} \cup R_{pref}, \tau \rangle$, is ground if no variables appearing in any of the probability logic rules in $R_{gen}$ or in any of the preference rules in $R_{pref}$

\begin{definition} Let $\Pi = \langle R_{gen} \cup R_{pref}, \tau \rangle$ be a ground probability answer set optimization program, $h$ be a probability answer set of $R_{gen}$(possibly partial), and $r$ be a probability preference rule in $R_{pref}$ of the form (\ref{rule:pref}). Then the satisfaction of a boolean combination, $C$, appearing in the $head(r)$ by $h$, denoted by $h \models C$, is defined inductively as follows:

\begin{itemize}

\item $h \models L:\mu$ iff  $\mu \leq_t h(L)$.

\item $h \models not\;L:\mu$ iff $\mu \nleq_t h(L)$ or $L$ is undefined in $h$.

\item $h \models C_1 \wedge C_2$ iff $h \models C_1$ and $h \models C_2$.

\item $h  \models C_1 \vee C_2$ iff $h \models C_1$ or $h \models C_2$.
\end{itemize}
Given $L_i:\mu_i$ and $not\;L_j:\mu_j$ appearing in  $body(r)$, the satisfaction of $body(r)$ by $h$, denoted by $h \models body(r)$, is defined inductively as follows:
\begin{itemize}
\item $h \models L_i:\mu_i$ iff $\mu_i \leq_t h(L_i)$

\item $h \models not\;L_j:\mu_j$ iff $\mu_j \nleq_t h(L_j)$ or $L_j$ is undefined in $h$.

\item $h \models body(r)$ iff $\forall(k+1 \leq i \leq m)$, $h \models L_i : \mu_i$ and $\forall(m+1 \leq j \leq n)$, $h \models not\; L_j : \mu_j$.
\end{itemize}

\end{definition}
The satisfaction of probability preference rules is defined as follows.

\begin{definition} Let $\Pi = \langle R_{gen} \cup R_{pref}, \tau \rangle$ be a ground probability answer set optimization program, $h$ be a probability answer set for $R_{gen}$, and $r$ be a probability preference rule in $R_{pref}$, and $C_i$ be a boolean combination in $head(r)$. Then, we define the following notions of satisfaction of $r$ by $h$:

\begin{itemize}
\item $h \models_{i} r$ iff $h \models body(r)$ and $h \models C_i$.

\item $h \models_{irr} r$ iff $h \models body(r)$ and $h$ does not satisfy any $C_i$ in $head(r)$.

\item $h \models_{irr} r$ iff $h$ does not satisfy $body(r)$.
\end{itemize}
\end{definition}
$h \models_{i} r$ means that the body of $r$ and the boolean combination $C_i$ that appearing in the head of $r$ is satisfied by $h$. However, $h \models_{irr} r$ means that $r$ is irrelevant (denoted by $irr$) to $h$, or, in other words, the probability preference rule $r$ is not satisfied by $h$, because either one of two reasons. Either because the body of $r$ and non of the boolean combinations that appearing in the head of $r$ are satisfied by $h$. Or because the body of $r$ is not satisfied by $h$.

\begin{definition} Let $\Pi = \langle R_{gen} \cup R_{pref}, \tau \rangle$ be a ground probability answer set optimization program, $h_1, h_2$ be two probability answer sets of $R_{gen}$, $r$ be a probability preference rule in $R_{pref}$, and $C_i$ be boolean combination appearing in $head(r)$. Then, $h_1$ is strictly preferred over $h_2$ w.r.t. $C_i$, denoted by $h_1 \succ_i h_2$, iff $h_1 \models C_i$ and $h_2 \nvDash C_i$ or $h_1 \models C_i$ and $h_2 \models C_i$ and one of the following holds:

\begin{itemize}

\item $C_i = L:\mu$ implies $h_1 \succ_i h_2$ iff $h_1(L) > h_2(L)$.

\item $C_i = not \; L:\mu$ implies $h_1 \succ_i h_2$ iff $h_1(L) < h_2(L)$ or $L$ is undefined in $h_1$ but defined in $h_2$.

\item $C_i = C_{i_1} \wedge C_{i_2}$ implies $h_1 \succ_i h_2$ iff there exists $t \in \{{i_1}, {i_2}\}$ such that $h_1 \succ_t h_2$ and for all other $t' \in \{{i_1}, {i_2}\}$, we have $h_1 \succeq_{t'} h_2$.

\item $C_i = C_{i_1} \vee C_{i_2}$ implies $h_1 \succ_i h_2$ iff there exists $t \in \{{i_1}, {i_2}\}$ such that $h_1 \succ_t h_2$ and for all other $t' \in \{{i_1}, {i_2}\}$, we have $h_1 \succeq_{t'} h_2$.

\end{itemize}
We say, $h_1$ and $h_2$ are equally preferred w.r.t. $C_i$, denoted by $h_1 =_{i} h_2$, iff $h_1 \nvDash C_i$ and $h_2 \nvDash C_i$ or $h_1 \models C_i$ and $h_2 \models C_i$ and one of the following holds:

\begin{itemize}

\item $C_i = L:\mu$ implies $h_1 =_{i} h_2$ iff $h_1(L) = h_2(L)$.

\item $C_i = not \; L:\mu$ implies $h_1 =_{i} h_2$  iff $h_1(L) = h_2(L)$ or $L$ is undefined in both $h_1$ and $h_2$.

\item $C_i = C_{i_1} \wedge C_{i_2}$ implies $h_1 =_{i} h_2$ iff
\[
\forall \: t \in \{{i_1}, {i_2}\}, \; h_1 =_{t} h_2
\]

\item $C_i = C_{i_1} \vee C_{i_2}$ implies $h_1 =_{i} h_2$ iff
\[
|\{h_1 \succeq_{t} h_2 \: | \: \forall \: t \in \{{i_1}, {i_2}\} \}| = | \{ h_2 \succeq_{t} h_1 \: | \: \forall \: t \in \{{i_1}, {i_2}\} \}|.
\]

\end{itemize}
We say, $h_1$ is at least as preferred as $h_2$ w.r.t. $C_i$, denoted by $h_1 \succeq_i h_2$, iff $h_1 \succ_i h_2$ or $h_1 =_i h_2$.
\label{def:compination}
\end{definition}

\begin{definition} Let $\Pi = \langle R_{gen} \cup R_{pref}, \tau \rangle$ be a ground probability answer set optimization program, $h_1, h_2$ be two probability answer sets of $R_{gen}$, $r$ be a probability preference rule in $R_{pref}$, and $C_l$ be boolean combination appearing in $head(r)$. Then, $h_1$ is strictly preferred over $h_2$ w.r.t. $r$, denoted by $h_1 \succ_r h_2$, iff one of the following holds:
\begin{itemize}
\item $h_1 \models_{i} r$ and $h_2 \models_{j} r$ and $i < j$, \\
where $i = \min \{l \; | \; h_1 \models_l r \}$ and $j = \min \{l \; | \; h_2 \models_l r \}$.

\item $h_1 \models_{i} r$ and $h_2 \models_{i} r$ and $h_1 \succ_i h_2$, \\
where $i = \min \{l \; | \; h_1 \models_l r \} = \min \{l \; | \; h_2 \models_l r \}$.

\item $h_1 \models_{i} r$ and $h_2 \models_{irr} r$.
\end{itemize}
We say, $h_1$ and $h_2$ are equally preferred w.r.t. $r$, denoted by $h_1 =_{r} h_2$, iff one of the following holds:
\begin{itemize}
\item $h_1 \models_{i}  r$ and $h_2 \models_{i} r$ and $h_1 =_i h_2$, \\
where $i = \min \{l \; | \; h_1 \models_l r \} = \min \{l \; | \; h_2 \models_l r \}$.
\item $h_1 \models_{irr}  r$ and $h_2 \models_{irr} r$.
\end{itemize}
We say, $h_1$ is at least as preferred as $h_2$ w.r.t. $r$, denoted by $h_1 \succeq_{r} h_2$, iff $h_1 \succ_{r} h_2$ or $h_1 =_{r} h_2$.
\label{def:pref-rule}
\end{definition}
The previous two definitions characterize how probability answer sets are ranked with respect to a boolean combination and with respect to a probability preference rule. Definition \ref{def:compination} presents the ranking of probability answer sets with respect to a boolean combination. But, Definition \ref{def:pref-rule} presents the ranking of probability answer sets with respect to a probability preference rule. The following definitions specify the ranking of probability answer sets according to a set of probability preference rules.

\begin{definition} [Pareto Preference] Let $\Pi = \langle R_{gen} \cup R_{pref}, \tau \rangle$ be a probability answer set optimization program and $h_1, h_2$ be probability answer sets of $R_{gen}$. Then, $h_1$ is (Pareto) preferred over $h_2$ w.r.t. $R_{pref}$, denoted by $h_1 \succ_{R_{pref}} h_2$, iff there exists at least one probability preference rule $r \in R_{pref}$ such that $h_1 \succ_{r} h_2$ and for every other rule $r' \in R_{pref}$, $h_1 \succeq_{r'} h_2$. We say, $h_1$ and $h_2$ are equally (Pareto) preferred w.r.t. $R_{pref}$, denoted by $h_1 =_{R_{pref}} h_2$, iff for all $r \in R_{pref}$, $h_1 =_{r} h_2$.
\end{definition}

\begin{definition} [Maximal Preference] Let $\Pi = \langle R_{gen} \cup R_{pref}, \tau \rangle$ be a probability answer set optimization program and $h_1, h_2$ be probability answer sets of $R_{gen}$. Then, $h_1$ is (Maximal) preferred over $h_2$ w.r.t. $R_{pref}$, denoted by $h_1 \succ_{R_{pref}} h_2$, iff
\[
|\{r \in R_{pref} | h_1 \succeq_{r} h_2\}| > |\{r \in R_{pref} | h_2 \succeq_{r} h_1\}|.
\]
We say, $h_1$ and $h_2$ are equally (Maximal) preferred w.r.t. $R_{pref}$, denoted by $h_1 =_{R_{pref}} h_2$, iff
\[
|\{r \in R_{pref} | h_1 \succeq_{r} h_2\}| = | \{r \in R_{pref} | h_2 \succeq_{r} h_1\}|.
\]
\end{definition}
It is worth noting that the Maximal preference definition is more {\em general} than the Pareto preference definition, since the Maximal preference relation {\em subsumes} the Pareto preference relation.

\section{Nurse Restoring with Probability Preferences Problem}

Nurse restoring problem is well-known scheduling problem in Operation Research \cite{Nurse}. In this section, we extend the nurse restoring problem to allow nurses to express their quantitative and qualitative preferences in terms of probability values over their choices, creating a new version of the nurse restoring problem called {\em Nurse Restoring with Probability Preference Problem}. We show that nurse restoring with probability preferences problem can be easily and intuitively represented and solved in the probability answer set optimization framework.

Nurse restoring with probability preferences problem is a multi-objective scheduling problem with several conflicting factors, like the hospitals views of the continuing insurance of sufficient nursing service at minimum cost and the nurses quantitative and qualitative preferences over working hours and days off, where the hospital management must resolve that conflict, since in any hospital's budget, the nursing service is one of its largest components. To accomplish the scheduling process, the nurse manger must collect information regarding the nursing service demands and the nurses quantitative and qualitative preferences over the available working hours. Hospitals typically employ nurses to work in shifts that cover the twenty four hours of the day, namely {\em early, day, late}, and {\em night} shifts with their obvious meanings.

The aim is to assign nurses to shifts over days, weeks, or months in order to provide a certain level of care in terms of nursing service whereas taking into consideration each individual nurse quantitative and qualitative preferences over shifts so that fairness and transparency are assured. Nurse preferences over shifts on a given day is given as a probability distribution over shifts on that day. Nurse restoring with probability preferences problem is formalized as given in the following example.

\begin{example} Assume that we have $n$ different nurses (denoted by $a_1, \ldots, a_n$) that need to be assigned to shifts among $k$ different shifts per a day (denoted by $s_1, \ldots, s_k$) for $m$ different days (denoted by $d_1, \ldots, d_m$) with the nurse manger demanding that each nurse is assigned exactly one shift per day and no two nurses are assigned the same shift on the same day. Each nurse prefers to work at certain shifts at certain days over other shifts in these certain days. Each nurse preferences over shifts per a day is represented as a probability distribution over shifts per that day. This nurse restoring with probability preferences problem can be represented as a probability answer set optimization program, $\Pi = \langle R_{gen} \cup R_{pref}, \tau \rangle$, where $\tau$ is any arbitrary assignments of probabilistic p-strategies and $R_{gen}$ is a set of disjunctive hybrid probability logic rules with probability answer set semantics of the form:
\[
\begin{array}{r}
service(a_i, s_1, d_j): \mu_{ij,1}  \vee  service(a_i, s_2,d_j): \mu_{ij,2} \vee \ldots \\ \vee  service(a_i, s_k,d_j): \mu_{ij,k}  \leftarrow
\\
inconsistent:1  \leftarrow not \; inconsistent:1,  \\ service(A, S, D):V, service(A', S,D):V', A \neq A'
\end{array}
\]
$\forall(1 \leq i \leq n)$ and $\forall(1 \leq j \leq m)$, where $V, V'$ are probability annotation variables act as place holders and, for any ($1 \leq l \leq k$), $service(a_i, s_l, d_j): \mu_{ij,l} $ represents that nurse $a_i$ prefers to service at shift $s_l$ in day $d_j$ with probability $\mu_{ij,l}$ (which is the nurse preference in servicing at the shift $s_l$ in the day $d_j$). The first disjunctive hybrid probability logic rule represents a nurse preferences over shifts per day while the second disjunctive hybrid probability logic rule represents the constraints that a nurse is assigned exactly one shift per day and one shift in a given day cannot be assigned to more than one nurse.
\label{ex:schedule}
\end{example}

The set of probability preference rules, $R_{pref}$, of the probability answer set optimization program representation of the nurse restoring with probability preferences problem consists $\forall(1 \leq i \leq n)$ and $\forall(1 \leq j \leq m)$ of the probability preference rule
\[
\begin{array}{r}
service(a_i, s_1, d_j): \mu_{ij,1}  \succ  service(a_i, s_2,d_j): \mu_{ij,2} \\ \succ \ldots \succ  service(a_i, s_k,d_j): \mu_{ij,k}  \leftarrow
\end{array}
\]
where $\forall (1 \leq i \leq n)$ and $\forall (1 \leq j \leq m)$, we have $\mu_{ij,1} \geq  \mu_{ij,2} \geq \ldots \geq  \mu_{ij,k}$.

However, the probability preference rules, $R_{pref}$, of the probability answer set optimization program representation of the nurse restoring with probability preferences problem can be easily and intuitively modified according to the nurses preferences in many and very flexible ways. For example, it can be the case that nurse $a$ is neutral regarding servicing at shifts $s_1$ and $s_2$ in a day $d$ with probability value $0.2$ each. This means that shifts $s_1$ and $s_2$ in day $d$ are equally preferred to nurse $a$. Hence, this situation can be represented in nurse $a$ probability preference rule in $R_{pref}$ as
\begin{equation*}
service(a, s_1, d): 0.2  \vee service(a, s_1, d): 0.2 \leftarrow
\end{equation*}
Moreover, although nurse $a$ is neutral regarding servicing at shifts $s_1$ and $s_2$ in day $d$ with probability value $0.2$ each, it can be the case that nurse $a$ has more appeal in servicing at shift $s_1$ over shift $s_2$ in day $d$. Therefore, this situation can be intuitively represented in nurse $a$ probability preference rule in $R_{pref}$ as
\begin{equation*}
service(a, s_1, d): 0.2  \succ service(a, s_1, d): 0.2 \leftarrow
\end{equation*}
Furthermore, it can be the case that each nurse has the preference of servicing at several shifts per a day with varying degrees of probability values. This also can be easily and intuitively accomplished by replacing the disjunctive hybrid probability logic rules in $R_{gen}$, of the probability answer set optimization program, $\Pi = \langle R_{gen} \cup R_{pref}, \tau \rangle$, representation of the nurse restoring with probability preferences problem by the following set of disjunctive hybrid probability logic rules:
\[
\begin{array}{r}
service(a_i, s_1, d_j,X): \mu_{ij,1}  \vee  service(a_i, s_2,d_j,X): \mu_{ij,2} \\ \vee \ldots \vee
service(a_i, s_k,d_j, X): \mu_{ij,k}  \leftarrow
\\
\\
inconsistent:1  \leftarrow not \; inconsistent:1,  \\ service(A, S, D, X):V, service(A', S,D,X):V', A \neq A'
\\
\\
inconsistent:1  \leftarrow not \; inconsistent:1, \\ service(A,S,D, X):V, service(A,S,D, X'):V', X \neq X'
\end{array}
\]
$\forall(1 \leq i \leq n)$ and $\forall(1 \leq j \leq m)$ and for all possible values of $X$, where the variable, $X$, is a dummy variable, where the number of values that the dummy variable, $X$, takes is equal to the number of shifts that a nurse is allowed to service per day. For example if the maximum number of shifts for a nurse to service per day is two, then the variable $X$ can be assigned to any two dummy values, e.g., $X = x$ and $X = y$. For all possible values of $X$, the first disjunctive hybrid probability logic rule assigns multiple shifts per day, $d_j$, to a nurse $a_i$. The last two disjunctive hybrid probability logic rules ensure that a shift per day is not assigned more than once to the same nurse.

Moreover, the disjunctive hybrid probability logic rules in, $R_{gen}$, allow multiple nurses to be assigned to the same shifts per a day, which can be necessary in situations where large number of patients are required to be serviced at given shifts per a day. In this case the hospital management may need to bound the number of allowable nurses per a shift per day. This also can be represented in the probability answer set optimization framework as a constraint using aggregate atoms in the style of the aggregate atoms presented in \cite{Recur-aggr}.

In addition to replacing the probability preference rules in $R_{pref}$, of the probability answer set optimization program, $\Pi = \langle R_{gen} \cup R_{pref}, \tau \rangle$, representation of the nurse restoring with probability preferences problem by the following probability preference rule $\forall(1 \leq i \leq n)$ and $\forall(1 \leq j \leq m)$ and for all possible values of $X$:
\[
\begin{array}{r}
service(a_i, s_1, d_j,X): \mu_{ij,1}  \succ  service(a_i, s_2,d_j,X): \mu_{ij,2} \\ \succ \ldots \succ service(a_i, s_k,d_j,X): \mu_{ij,k}  \leftarrow
\end{array}
\]
where $\forall (1 \leq i \leq n)$ and $\forall (1 \leq j \leq m)$, we have $\mu_{ij,1} \geq  \mu_{ij,2} \geq \ldots \geq  \mu_{ij,k}$. This shows in general that probability answer set optimization programs can be intuitively and flexibly used to represent and reason in the presence of both probability quantitative preferences and qualitative preferences. This is illuminated by the following instance of the nurse restoring with probability preferences problem described below.

\begin{example} Assume that the nurse manger wants to schedule the nursing service for the Saturday and Sunday of this week. However, three nurses, Jeen, Lily, and Lucci are available over the weekends of this week. Jeen, Lily, and Lucci probability quantitative and qualitative preferences over shifts per this Saturday and Sunday are given as described below. In addition, each of the nurses requires to be assigned exactly one shift per day and the nurse manger requires that no two nurses are assigned the same shift on the same day.

This instance of the nurse restoring with probability preferences problem can be represented as an instance of the probability answer set optimization program, $\Pi = \langle R_{gen} \cup R_{pref}, \tau \rangle$, presented in Example \ref{ex:schedule}, as a probability answer set optimization program, $\Pi' = \langle R'_{gen} \cup R'_{pref}, \tau \rangle$, where in addition to the last disjunctive hybrid probability logic rule in $R_{gen}$ of $\Pi$ described in Example \ref{ex:schedule}, $R'_{gen}$ also contains the following disjunctive hybrid probability logic rules:
{\small
\[
\begin{array}{l}
service(jeen, early, sat):0.8  \vee   service(jeen, day, sat):0.4  \leftarrow  \\
service(lily, day, sat):0.6  \vee   service(lily, late, sat):0.2  \leftarrow  \\
service(lucci, late, sat):0.3  \vee   service(lucci, night, sat):0.7  \leftarrow  \\
service(lucci, night, sun):0.7  \vee   service(lucci, early, sun):0.5  \leftarrow
\end{array}
\]
}
In addition, $R'_{pref}$, contains the probability preference rules:
{\small
\[
\begin{array}{l}
r_1: service(jeen, early, sat):0.8 \succ  service(jeen, day, sat):0.4  \leftarrow  \\
r_2: service(lily, day, sat):0.6   \succ   service(lily, late, sat):0.2  \leftarrow  \\
r_3: service(lucci, night, sat):0.7  \succ   service(lucci, late, sat):0.3   \leftarrow  \\
r_4: service(lucci, night, sun):0.7   \succ  service(lucci, early, sun):0.5  \leftarrow
\end{array}
\]
}
The generator rules, $R'_{gen}$, of the probability answer set optimization program, $\Pi'$, has eight probability answer sets that are:
{\small
\[
\begin{array}{l}
h_1 =  \{
service(jeen,day,sat):0.4, service(lily,late,sat):0.2, \\
service(lucci,night,sat):0.7, service(lucci,early,sun):0.5\}
\end{array}
\]
\[
\begin{array}{l}
h_2 =  \{
service(jeen,early,sat):0.8,  service(lily,late,sat):0.2, \\
service(lucci,night,sat):0.7,  service(lucci,early,sun):0.5
\}
\end{array}
\]
\[
\begin{array}{l}
h_3 =   \{
service(jeen,day,sat):0.4,   service(lily,late,sat):0.2, \\
   service(lucci,night,sat):0.7,    service(lucci,night,sun):0.7
\}
\end{array}
\]
\[
\begin{array}{l}
h_4 =   \{
service(jeen,early,sat):0.8,  service(lily,late,sat):0.2, \\
   service(lucci,night,sat):0.7,  service(lucci,night,sun):0.7
\}
\end{array}
\]
\[
\begin{array}{l}
h_5 =   \{
service(jeen,early,sat):0.8,  service(lily,day,sat):0.6, \\
   service(lucci,late,sat):0.3,   service(lucci,early,sun):0.5
\}
\end{array}
\]
\[
\begin{array}{l}
h_6 =   \{
service(jeen,early,sat):0.8,  service(lily,day,sat):0.6, \\
   service(lucci,late,sat):0.3,  service(lucci,night,sun):0.7
\}
\end{array}
\]
\[
\begin{array}{l}
h_7 =   \{
service(jeen,early,sat):0.8,     service(lily,day,sat):0.6, \\
   service(lucci,night,sat):0.7,     service(lucci,early,sun):0.5
\}
\end{array}
\]
\[
\begin{array}{l}
h_8 =   \{
service(jeen,early,sat):0.8,   service(lily,day,sat):0.6, \\
   service(lucci,night,sat):0.7,  service(lucci,night,sun):0.7
\}
\end{array}
\]
}
We can easily verify that
\[
\begin{array}{llll}
h_1 \models_{2} r_1, & \; h_1 \models_{2} r_2, & \;  h_1 \models_{1} r_3, & \;  h_1 \models_{2} r_4 \\
h_2 \models_{1} r_1, & \; h_2 \models_{2} r_2, & \;  h_2 \models_{1} r_3, & \;  h_2 \models_{2} r_4 \\
h_3 \models_{2} r_1, & \; h_3 \models_{2} r_2, & \;  h_3 \models_{1} r_3, & \;  h_3 \models_{1} r_4 \\
h_4 \models_{1} r_1, & \; h_4 \models_{2} r_2, & \;  h_4 \models_{1} r_3, & \;  h_4 \models_{1} r_4 \\
h_5 \models_{1} r_1, & \; h_5 \models_{1} r_2, & \;  h_5 \models_{2} r_3, & \;  h_5 \models_{2} r_4 \\
h_6 \models_{1} r_1, & \; h_6 \models_{1} r_2, & \;  h_6 \models_{2} r_3, & \;  h_6 \models_{1} r_4 \\
h_7 \models_{1} r_1, & \; h_7 \models_{1} r_2, & \;  h_7 \models_{1} r_3, & \;  h_7 \models_{2} r_4 \\
h_8 \models_{1} r_1, & \; h_8 \models_{1} r_2, & \;  h_8 \models_{1} r_3, & \;  h_8 \models_{1} r_4
\end{array}
\]
It can be seen that the top (Pareto and Maximal) preferred probability answer set is $h_8$ and the least (Maximal) preferred probability answer set is $h_1$. The probability answer sets $h_2$, $h_3$, and $h_5$ are equally (Maximal) preferred. In addition, the probability answer sets $h_4$, $h_6$, and $h_7$ are equally (Maximal) preferred. However, any of the probability answer sets $h_4$, $h_6$, and $h_7$ is (Maximal) preferred over any of the probability answer sets $h_2$, $h_3$, and $h_5$. Therefore, the ranking of the probability answer sets, with respect to $R_{pref}'$, according to the Maximal preference is given as \[h_8 \succ h_4  = h_6 = h_7 \succ h_2 = h_3 = h_5 \succ h_1\]
\end{example}

\section{Properties}

In this section we prove that the probability answer set optimization programs syntax and semantics naturally generalize and subsume the classical answer set optimization programs syntax and semantics \cite{ASO} under the Pareto preference relation, since there is no notion of Maximal preference relation has been defined for the classical answer set optimization programs.

A classical answer set optimization program, $\Pi^c$, consists of two separate classical logic programs which are a classical answer set program, $R^c_{gen}$, and a classical preference program, $R^c_{pref}$ \cite{ASO}. The classical answer set program, $R^c_{gen}$, is used to generate the classical answer sets, however, the classical preference program, $R^c_{pref}$, defines classical context-dependant preferences that are used to form a preference ordering among the classical answer sets of $R^c_{gen}$.

Every classical answer set optimization program, $\Pi^c = R^c_{gen} \cup R^c_{pref}$, is represented as a probability answer set optimization program, $\Pi = \langle R_{gen} \cup R_{pref}, \tau \rangle$, where all probability annotations appearing in every probability logic rule in $R_{gen}$ and all probability annotations appearing in every probability preference rule in $R_{pref}$ is equal to $[1,1]$, which means the truth value {\em true}, and $\tau$ is any arbitrary mapping $\tau: {\cal B_L} \rightarrow S_{disj}$. For example, a classical answer set optimization program, $\Pi^c = R^c_{gen} \cup R^c_{pref}$, which is represented by the probability answer set optimization program, $\Pi = \langle R_{gen} \cup R_{pref}, \tau \rangle$, contains the classical logic rule
\begin{eqnarray*}
a_1 \; \vee \ldots \vee \; a_k \leftarrow  a_{k+1}, \ldots, a_m, not\; a_{m+1},
\ldots, not\;a_{n}
\end{eqnarray*}
in $R^c_{gen}$, where $\forall (1 \leq i \leq n)$, $a_i$ is an atom, iff
\begin{eqnarray*}
a_1:[1,1] \; \vee \ldots \vee \; a_k:[1,1] \leftarrow  a_{k+1}:[1,1], \ldots, a_m:[1,1], \\ not\; a_{m+1}:[1,1],
\ldots, not\;a_{n}:[1,1]
\end{eqnarray*}
is contained in $R_{gen}$. It is worth noting that the syntax and semantics of this class of probability answer set programs are equivalent to the syntax and semantics of the classical answer set programs \cite{Saad_DHPP,Saad_EHPP}. Moreover, the classical preference rule
\begin{eqnarray*}
C_1 \succ C_2 \succ \ldots \succ C_k \leftarrow l_{k+1},\ldots, l_m,
not\; l_{m+1},\ldots, not\;l_{n}
\end{eqnarray*}
belongs to $R^c_{pref}$, where $l_{k+1}, \ldots, l_{n}$ are literals and $C_1, C_2, \ldots, C_k$ are boolean combinations over a set of literals, iff
\[
\begin{array}{r}
C_1 \succ C_2 \succ \ldots \succ C_k \leftarrow l_{k+1}:[1,1],\ldots, l_m:[1,1], \\
not\; l_{m+1}:[1,1],\ldots, not\;l_{n}:[1,1]
\end{array}
\]
belongs to $R_{pref}$ and $C_1, C_2, \ldots, C_k$ are the same boolean combinations as in the classical preference rule in addition to every literal appearing in $C_1, C_2, \ldots, C_k$ is annotated with the probability annotation $[1,1]$.

Assuming that \cite{ASO} assigns the lowest rank to the classical answer sets that do not satisfy either the body of a classical preference rule or the body of a classical preference and any of the boolean combinations appearing in the head of a classical preference rule, the following theorems prove that the syntax and semantics of the probability answer set optimization programs subsume the syntax and semantics of the classical answer set optimization programs \cite{ASO}.

\begin{theorem} Let $\Pi = \langle R_{gen} \cup R_{pref}, \tau \rangle$ be a probability answer set optimization program equivalent to a classical answer set optimization program, $\Pi^c = R^c_{gen} \cup R^c_{pref}$. Then, the preference ordering of the probability answer sets of $R_{gen}$ w.r.t. $R_{pref}$ coincides with the preference ordering of the classical answer sets of $R^c_{gen}$ w.r.t. $R^c_{pref}$.
\label{thm:1}
\end{theorem}

\begin{theorem} Let $\Pi = \langle R_{gen} \cup R_{pref}, \tau \rangle$ be a probability answer set optimization program equivalent to a classical answer set optimization program, $\Pi^c = R^c_{gen} \cup R^c_{pref}$. A probability answer set $h$ of $R_{gen}$ is Pareto preferred probability answer set w.r.t. $R_{pref}$ iff a classical answer set $I$ of $R^c_{gen}$, equivalent to $h$, is Pareto preferred classical answer set w.r.t. $R^c_{pref}$.
\label{thm:2}
\end{theorem}

\section{Conclusions and Related Work}

Syntax and semantics of a unified logical framework for representing and reasoning about both probability quantitative and qualitative preferences, called probability answer set optimization programs has been developed. The presented logical framework is necessary to allow representing and reasoning in the presence of both probability quantitative and qualitative preferences across probability answer sets, which in turn, allows the ranking of the probability answer sets from the top preferred probability answer set to the least preferred probability answer set, where the top preferred probability answer set is the one that is most desirable. Probability answer set optimization framework generalizes and modifies the classical answer set optimization programs proposed in \cite{ASO}. We have shown the application of probability answer set optimization programs to the nurse restoring with probability preferences problem. Furthermore, we described an implementation for probability answer set optimization programs using probability answer set programming. To the best of our knowledge, this development is the first to consider a logical framework for reasoning about probability quantitative preferences, in general, and reasoning about both quantitative and qualitative preferences in particular.

On the other hand, qualitative preferences were introduced in classical answer set programming in various forms. In \cite{Schaub-Comp}, qualitative preferences are defined among the rules of classical logic programs, whereas qualitative preferences among the literals described by the classical logic programs are introduced in \cite{Sakama}. Classical answer set optimization \cite{ASO} and classical logic programs with ordered disjunctions \cite{LPOD} are two classical answer set programming based qualitative preference handling approaches, where context-dependant qualitative preferences are defined among the literals specified by the classical logic programs. Application-dependant qualitative preference handling approaches for planning were presented in \cite{Son-Pref,Schaub-Pref07}, where qualitative preferences among actions, states, and trajectories are defined, which are based on temporal logic. The major difference between \cite{Son-Pref,Schaub-Pref07} and \cite{ASO,LPOD} is that the former are specifically developed for planning, but the latter are application-independent.

Contrary to the existing approaches for reasoning about qualitative preferences in classical answer set programming, where qualitative preference relations are specified among rules and literals in one classical logic program, a classical answer set optimization program consists of two separate classical logic programs; a classical answer set program and a qualitative preference program \cite{ASO}. The classical answer set program is used to generate the classical answer sets and the qualitative preference program defines context-dependant qualitative preferences that are used to form a qualitative preference ordering among the classical answer sets generated by the classical answer set program.

Similar to \cite{ASO}, probability answer set optimization programs presented in this paper distinguish between probability answer sets generation and probability preference based probability answer sets evaluation, which has several advantages. In particular, the set of probability preference rules, in a probability answer set optimization program, is specified independently from the type of probability logic rules used to generate the probability answer sets in the probability answer set optimization program, which makes preference elicitation easier and the whole approach more intuitive and easy to use in practice. In addition, more expressive forms of probability preferences can be represented in probability answer set optimization programs, since they allow several forms of boolean combinations in the heads of the probability preference rules.

In \cite{Saad_ASOG}, the classical answer set optimization programs have been extended to allow classical aggregate preferences. The introduction of classical aggregate preferences to classical answer set optimization programs have made the encoding of multi-objectives optimization problems and Nash equilibrium strategic games more intuitive and easy. The syntax and semantics of the classical answer set optimization programs with classical aggregate preferences were based on the syntax and semantics of classical answer set optimization programs \cite{ASO} and classical aggregates in classical answer set programming \cite{Recur-aggr}. It has been shown in \cite{Saad_ASOG} that the syntax and semantics of classical answer set optimization programs with classical aggregate preferences subsumes the syntax and semantics of classical answer set optimization programs described in \cite{ASO}.

\bibliographystyle{named}
\bibliography{Saad13LPP}

\end{document}